\documentclass[runningheads,usenames,dvipsnames]{llncs}
\usepackage{graphicx}
\usepackage{amsmath,amssymb} 
\usepackage{color}
\usepackage{booktabs,verbatim} 
\usepackage[dvipsnames]{xcolor}

\makeatletter
\newcommand{\eqnum}{\refstepcounter{equation}\textup{\tagform@{\theequation}}}
\makeatother
\begin{document}
\pagestyle{headings}
\mainmatter

\def\ACCV18SubNumber{165}  

\title{Depth Reconstruction of Translucent Objects from a Single Time-of-Flight Camera using Deep Residual Networks} 
\titlerunning{ACCV-18 submission ID \ACCV18SubNumber}
\authorrunning{ACCV-18 submission ID \ACCV18SubNumber}

\author{Seongjong Song and Hyunjung Shim}
\institute{School of Integrated Technology, Yonsei University, South Korea}

\maketitle

\begin{abstract}
We propose a novel approach to recovering the translucent objects from a single time-of-flight (ToF) depth camera using deep residual networks. When recording the translucent objects using the ToF depth camera, their depth values are severely contaminated due to complex light interactions with the surrounding environment. While existing methods suggested new capture systems or developed the depth distortion models, their solutions were less practical because of strict assumptions or heavy computational complexity. In this paper, we adopt the deep residual networks for modeling the ToF depth distortion caused by translucency. To fully utilize both the local and semantic information of objects, multi-scale patches are used to predict the depth value. Based on the quantitative and qualitative evaluation on our benchmark database, we show the effectiveness and robustness of the proposed algorithm.

\end{abstract}

\section{Introduction}

Depth cameras are widely used in various applications including augmented reality, game, human-computer interaction and scene understanding. Owing to their real-time performance,  portability, and the reasonable price, depth cameras are commercially succeeded, even built in the smartphones. (e.g., iPhone X)  Existing depth sensing technologies utilize the active light source projected onto the target object, and analyze their light reflection to acquire the depth value of 3-D point. 

However, the appearance of translucent object is determined by the complex light interactions associated with the light refraction and transmission. Consequently, when we capture the translucent object using a commercial depth camera, the resultant depth map presents significant errors. Although this issue is well-known and also considered to be critical in research community, it remains unsolved because this error is closely involved with understanding the surroundings of target objects. Lately, several approaches attempt to address this problem by either 1) utilizing the controlled environment \cite{clark1997using,kim2017acquiring,tanaka2016recovering,hullin2008fluorescent}, 2) developing the empirical model of depth error \cite{fujimura2018photometric,maeno2013light}, or 3) exploiting the context information from RGB images \cite{wang2012glass,seib2017friend}. Controlling the capture environment formulates the problem of reconstructing the translucent object well-posed. Although they produce high quality geometric models, these solutions are limited to laboratory experiment, not suitable for practical applications. Later, empirical error models are introduced for releasing the capture conditions. However, their performance is quite limited to the specific types of objects, incapable of handling various shapes and materials in translucent objects. Inspired by the human perceptual ability that recognizes the translucent objects, several studies extract the context information from RGB images and utilize it for understanding the translucent objects. Yet, these approaches do not cover the problem of depth reconstruction.  

In this paper, we propose a learning-based approach to compensating the depth distortion in translucent objects using a single time-of-flight camera. We utilize both the foreground depth map and background depth map to correct the depth distortion as inputs and recover the correct depth map for the translucent object. Compared to several approaches utilizing RGB images, our algorithm is robust to harsh lighting conditions or dark environment. Also, it is worthwhile noting that the proposed algorithm is a purely data driven approach. That means, we do not require physical constraints \cite{shim2016recovering,kim2017robust} or the controlled environment for developing the model. Consequently, while existing work should be reformulated even with the slight modifications on their assumptions, the proposed framework is extendable to different conditions or scenarios as long as the additional datasets are available.

More specifically, we develop deep convolutional networks for recovering translucent objects from depth maps. Our network architecture is inspired by deep residual network \cite{he2016deep}, or ResNet, which has been successfully adopted to object classification and image restoration problems \cite{nah2016deep}. On top of ResNet architecture, our model simultaneously processes multi-scale patches from two input images. In this way, we intend to cope both the local characteristics (i.e., small scale patch) and semantic information of target objects (i.e., large scale patch). As a result, the proposed algorithm improves the accuracy of reconstructing translucent objects both quantitatively and qualitatively. Particularly, we show that our algorithm is robust against various levels of sensor noise, which is inevitable in most of time-of-flight depth sensor due to short exposures and the limited amount of emitted light energy. To the best of our knowledge, the proposed algorithm is the first attempt to solving the 3D reconstruction of translucent objects using deep neural networks. We believe our work can serve an important baseline of future work. For that, we will make our database and code publicly available upon the acceptance of paper.

\section{Related Work}

\textbf{Depth Acquisition of Transparent Objects.} Recovering 3-D transparent objects is known to be a challenging research problem in computer vision, and numerous techniques have been proposed to address this problem.
Previous techniques utilized the laser scan with polarization \cite{clark1997using} or with a fluorescent liquid \cite{hullin2008fluorescent} for the accurate reconstruction of translucent objects. Tanaka et al. \cite{tanaka2016recovering}  recovered the 3D shape of transparent objects utilizing a known refractive index and the images captured under controlling the background patterns. Kim et al. \cite{kim2017acquiring} analyzed the images recorded by projecting several background patterns, and then reconstructed the shape of axially-symmetric transparent objects. Despite of the impressive quality, their capture systems rely on the controlled environment,  unlikely applyable to practical applications. 

Several approaches observe that the depth distortions caused by transparent objects provide meaningful clues for understanding their shapes, even in practical situations. Maeno et al. \cite{maeno2013light} analyzed the light field distortion caused by transparent objects (e.g, a glass) and used it for recognizing the object. However, detecting depth distortion features heavily depends on the type of background; for instance, a textureless background or a scene with repeating patterns degrades the detectability. Torres-G\'omez and Mayol-Cuevas \cite{torres2014recognition} segmented and roughly reconstructed transparent object from multiple color images by carefully stitching hand-crafted features for translucent objects. Yet, their algorithm is limited to a glass object with a spatially smooth surface.

When acquiring translucent objects using the structured light based depth sensors, those target objects often appear invalid, shown as empty holes in the depth map. From this observation, several techniques detect the holes in the depth map and evaluate those holes using its RGB image to classify or localize the translucent objects. That is, Wang et al. \cite{wang2012glass} sequentially applied traditional image processing algorithms for transparent object localization while Lysenkov et al. \cite{lysenkov2013recognition} adopted the template matching for both localization and pose estimation. Most recently, owing to the rapid progress in machine learning algorithms, the correspondence problem in depth estimation is being addressed by learning based approach \cite{fanello2016patent,hartmann2017learned}. Seib et al. \cite{seib2017friend} extended learning-based approach to an end-to-end framework using Convolutional Neural Network (CNN). They handled depth maps containing transparent objects, yet focusing on classifying and localizing predefined objects exploiting holes of depth image. 
Yet, it is important to note that our goal is to recover the depth distortion of translucent object, captured by the ToF sensors;  depth distortions in ToF sensors produce incorrect depth values instead of holes.


\textbf{Time-of-flight Multipath Inference (MPI) Correction. }
In any case of depth acquisition, the recorded signal might be the result of combining multiple reflected signals from the source, each travels from different paths. This is namely the multi-path inference (MPI), which causes significant depth errors in a concave object or corner of the scene (indirect bounces), in mildly translucent material such as skin or wax (subsurface scattering), or along dense participating media. Alleviating those errors is an active research topic for various imaging systems \cite{feigin2015resolving,naik2015light,fujimura2018photometric}. In case of the ToF depth sensors, MPI leads to the depth value farther than ground truth because the longer traveling path results in the larger depth. Marco et al. \cite{marco2017deeptof} corrected MPI using a single ToF depth image and without any additional information, by learning from the simulated data, such as simulated indirect bounces and global illumination. Naik et al. \cite{naik2015light} used high-frequency illumination patterns to resolve ambiguities in multiple possible travel paths, reducing errors caused by sub-surface scattering.

Compared with the MPI problem, recovering transparent or translucent objects should handle a severe case of subsurface scattering. Because translucent objects present the light scattering as well as transmission, the amount of depth distortions is considerably larger than the distortions caused by MPI. 

Shim and Lee \cite{shim2016recovering} also reported the same analysis, and suggested the depth distortion model based on a ToF depth sensing principle for reconstructing translucent objects. They showed that their model is effective to restore the 3D shape of translucent objects assuming a known background and ignoring the effect of light refraction. Later, Kim and Shim \cite{kim2017robust} extended the idea of \cite{shim2016recovering} and improved the performance by integrating user interactions. Still, these models assume no refraction, limiting the pose of target object being frontal and the shape being planar. Such assumptions are too strict for practical applications.


\textbf{Residual Networks.}
Since AlexNet \cite{krizhevsky2012imagenet} was announced with outstanding performance in image classification, CNN has been successfully applied in visual recognition tasks. While increasing the number of stacked layers (network depth) of CNN is expected to improve high level feature extraction and boost the performance, the tradeoff is to hinder the training process; it can lead to the gradient vanishing during backpropagation. He et al. \cite{he2016deep} overcame this drawback by stacking residual blocks, while each block includes an identity mapping to directly link between its input and output. Their invention is called a ResNet architecture, and it is widely applied not only to a visual recognition task or segmentation, but also in more complicated tasks including image restoration \cite{nah2016deep,iizuka2017globally}, style/domain transfer \cite{gatys2016image,liao2017visual}, and depth estimation \cite{li2017single,cao2017estimating,lee2018single}.


\begin{figure*}[!t]
    \centerline{\includegraphics[width=\textwidth]{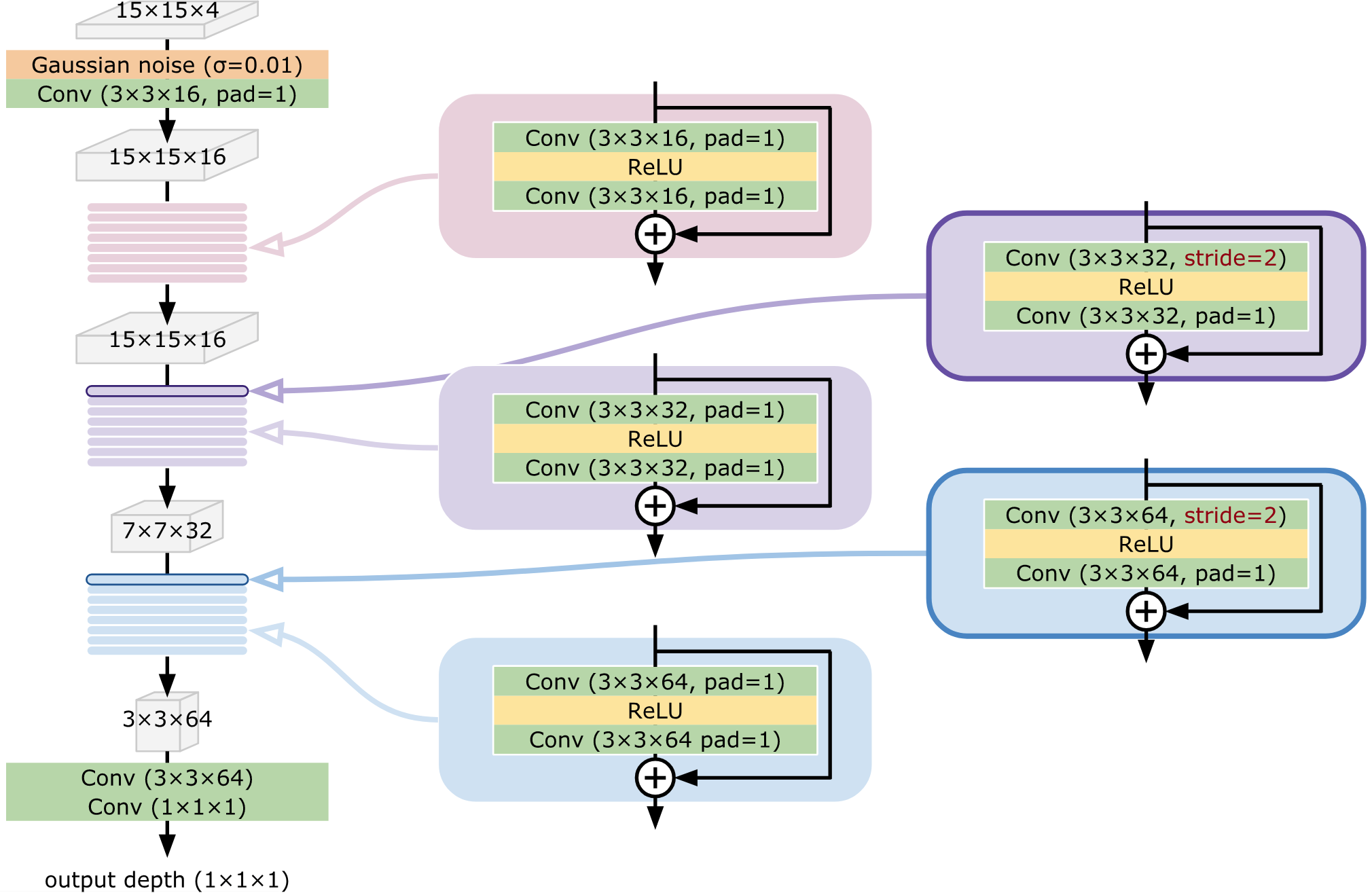}}
    \caption{Illustration of proposed network architecture. Main part of the model consists of 3 groups of 8 stacked residual blocks. Bordered blocks ({\it on the top of 2nd and 3rd group}) exceptionally have a convolution layer of \texttt{pad\,$=0$} and \texttt{stride\,$=2$} to downsample the input size by half
    }
    \label{fig1}
\end{figure*}

\section{Multi-scale Patch based Residual Networks}

To solve the depth reconstruction problem, we introduce three important ideas for developing the network. 

First of all, we formulate the depth reconstruction problem by the patch-based regression. Several existing techniques \cite{li2017single,cao2017estimating} adopt deep neural networks for estimating the depth map from RGB images by interpreting the depth estimation as a classification problem via depth binning. Although this leads the depth reconstruction problem being simplified, their results inherently exhibit the quantization errors, and always require post-processing to generate continuous depth maps. To prevent this issue, our network is trained to directly map an input patch to a single depth value. At the last convolutional layer, the patch is reduced to $1 \times 1 \times 1$, and this corresponds to an estimated depth value at the center of patch. 
Also, our patch-based approach is advantageous for training using the limited database. While the image-based approach learns the overall structures and semantic information better, the corresponding network should be accompanied by much larger amount of model parameters; it requires to establish much larger training database.  

Secondly, we leverage multi-scale patches (i.e., small and large scale patches) for estimating the depth value. For the image generation task, considering both semantic consistency and local details is difficult yet important problem. One of the means to solve this problem is the multi-scale approach, used by Iizuka et al. \cite{iizuka2017globally}. Their model includes two discriminator networks, each processing a different image patch of two different resolutions. Influenced by them, we compose input patches by the concatenation of original image and $1/4$ resolution image. By considering the two different scales during training, we effectively increase the receptive field size while maintaining the capability of representing details. Although we add the full-scale image into the input patches, it is different from the image based approach, which transforms the image to the depth map. Because both the full-scale image and the $1/4$ scale image are reduced and eventually reach a single depth value after passing those of residual blocks, the number of parameters is still tractable, and is able to be trained on a relatively small size of database.

Finally, we remove the batch normalization (BN) layers as opposed to the original ResNet architecture. BN was first introduced by Ioffe and Szegedy \cite{ioffe2015batch} to stabilize the training and accelerate the convergence of loss. However, unlike conventional belief of BN reducing internal covariance shift, most recent researches \cite{santurkar2018does,kohler2018towards} show that the role of BN for accelerating training is still not clearly evidenced.  
Furthermore in our problem, restoring the absolute depth value is important because the estimated depth value is later placed with surrounding depth values in the final depth map. If the BN layer is adapted into the network, it forces to normalize the data distribution from each batch during the training phase. As a result, during the test phase, the input patch maps to the relative depth value because of the normalization effect. Motivated by \cite{nah2016deep}, we successfully stabilize the training without the BN, by relatively reducing the training batch size and learning rate.

\subsection{Acquiring the Database}

\begin{figure}[tb]
    \centering
    \begin{minipage}[b]{.13\linewidth}
      \centering
      \centerline{\includegraphics[width=1.6cm]{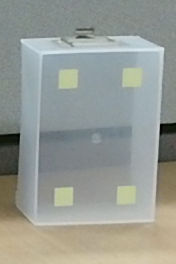}}
    \end{minipage}
    \begin{minipage}[b]{.13\linewidth}
      \centering
      \centerline{\includegraphics[width=1.6cm]{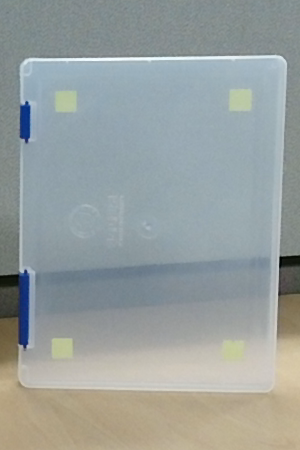}}
    \end{minipage}
    \begin{minipage}[b]{.13\linewidth}
      \centering
      \centerline{\includegraphics[width=1.6cm]{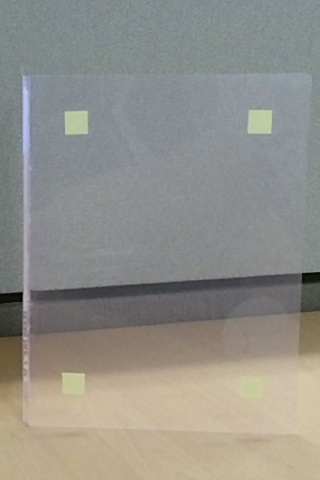}}
    \end{minipage}
    \begin{minipage}[b]{.13\linewidth}
      \centering
      \centerline{\includegraphics[width=1.6cm]{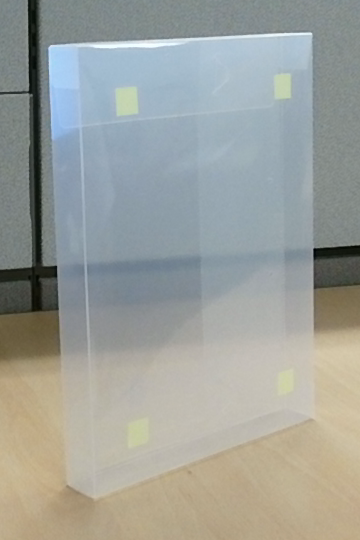}}
    \end{minipage}
    \begin{minipage}[b]{.13\linewidth}
      \centering
      \centerline{\includegraphics[width=1.6cm]{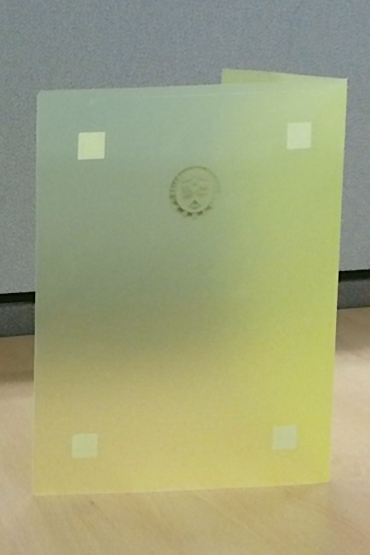}}
    \end{minipage}
    \begin{minipage}[b]{.13\linewidth}
      \centering
      \centerline{\includegraphics[width=1.6cm]{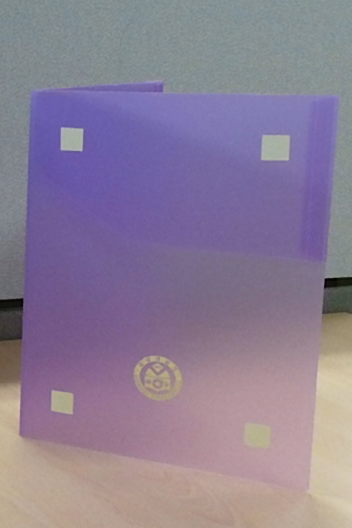}}
    \end{minipage}
    
    \hfill\vline\hfill
    
    \centering
    \begin{minipage}[b]{.19\linewidth}
      \centering
      \centerline{\includegraphics[height=2.2cm]{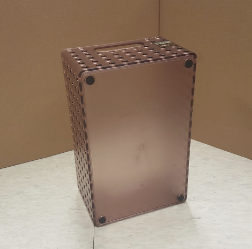}}
    \end{minipage}
    \begin{minipage}[b]{.19\linewidth}
      \centering
      \centerline{\includegraphics[height=2.2cm]{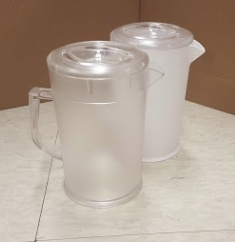}}
    \end{minipage}
    \begin{minipage}[b]{.19\linewidth}
      \centering
      \centerline{\includegraphics[height=2.2cm]{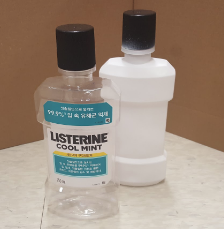}}
    \end{minipage}
    \begin{minipage}[b]{.19\linewidth}
      \centering
      \centerline{\includegraphics[height=2.2cm]{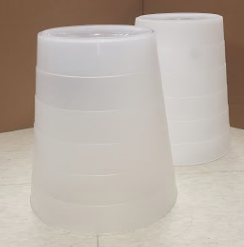}}
    \end{minipage}
    
    \caption{Target translucent objects. Top row: six training objects with markers. Bottom row: one flat object and three round objects for testing}
    \label{fig2}
\end{figure}

For supervised learning, it is necessary to obtain ground truth depth data. However, acquiring the ground truth itself can be another research problem in case of translucent objects. Even if the prior for object shape is known by any means, registering the prior shape to the depth map is prone to errors due to severe depth distortions caused by translucency. To bypass the registration issue, we use extra makers to simultaneously register and recover the ground truth shape of training objects, as shown in Fig.~\ref{fig2}.

To acquire the ground truth depth data regardless of optical characteristics, we utilize the objects with known shapes, particularly planar objects. Four thin and opaque markers are attached to four corners of rectangular objects, 3 cm apart from each boundary. When a raw depth map is captured via ToF camera, we extract depth values at opaque markers on the object and use them as undistorted depth values. We assume the object surface as the projective transformation of a rectangle. Utilizing the width and height of the object, and position of four markers on the image, a rectangular mask is able to be fitted into the captured image via projective transform. Note that we can identify the width and height of the target object from the raw depth map because its x and y coordinates are still valid. Four true depth values on the center of each marker are interpolated and extrapolated to fit a plane and fill the depth mask.

To evaluate how much our algorithm is sensitive to the shape of objects, we also collect the ground truth depth maps of three different round objects. For each test object, we prepare two identical objects; serving one for the ground truth depth map and the other for distorted depth map. For acquiring the ground truth depth map, we apply white matt spray to coat its surface. For the evaluation, we generate the object mask by thresholding the depth difference between background and ground truth depth map. It is important to emphasize that this mask does not input to our network, thus the mask is not required for testing phase. Instead, this mask is utilized only for preparing the training dataset, visualizing the result, and conducting the quantitative evaluation.

We further divide recorded depth maps into patches for constructing the training dataset. Each training sample consists of 4 $15 \times 15$ images. Among those four images, two images ($A$) represent the difference between raw depth and background depth. The other two ($B$) consist of masked raw depth, whose pixel value is 0 outside the object. Then, each of two images is from either original or $1/4$ resolution patch, namely $A_1$, $A_{1/4}$, $B_1$ and $B_{1/4}$. To fit those four images into $15 \times 15$ resolution, we choose the nearest neighbor interpolation; the $1/4$ resolution patch is obtained by sampling pixels with stride of 4. Then, the 2D patches are concatenated to fit into the $15 \times 15 \times 4$ tensor, which is our network input.

\subsection{Training Details}

We capture depth images of 10 translucent objects with Kinect v2. With data augmentation of horizontal flip, approximately 350k patches from 48 recorded depth maps (corresponding to 10 objects with various background conditions and poses) are used for training. The proposed network is trained on three stages: 10 epochs with learning rate $\eta=0.0003$, 20 epochs in $\eta=0.0001$, and 20 epochs in $\eta=0.000033$. The learning rate greater than $0.0003$ triggers sporadic sparks of loss and leads to unstable training. We use RMSProp \cite{tieleman2012lecture} with the momentum of $0.5$ and a smooth $L1$ loss as the objective function, and training batch size is $4$ through the entire training procedure.

We train our network using six planar objects attached with opaque markers. The remaining planar objects and three additional round objects are used for testing. 
For the evaluation, a $3 \times 3$ median filter is applied as post-processing.

\section{Experimental Results}

\subsection{Baseline models}

We evaluate the depth reconstruction accuracy of proposed model compared with that of Shim and Lee \cite{shim2016recovering}. Similar to our model, their approach also compensates depth distortions of translucent object under a known background. Nonetheless, their model is built upon the strong assumptions on the orientation and the properties of target object; the object should have a frontal pose and form a thin planar surface. Because their model is formulated by only considering the light transmission, these restrictions must be satisfied for their ideal performance.

In addition to the evaluation with the competitor, we investigate the influence of multi-scale patch and exclusion of batch normalization. To analyze the role of multi-scale patch, a network with identical structure as proposed is trained from scratch, using only the patch consists of full resolution. To observe the effect of batch normalization, another network is also trained under identical condition, with the structure modification of added batch normalization layers very next to each convolution layer in all 24 residual blocks.

\begin{figure*}[!tb]

\hspace*{\fill}
\begin{minipage}[b]{.16\linewidth}
  \centering
  \centerline{(a) Raw depth}\medskip
  \centerline{\includegraphics[trim=30px 10px 70px 40px, clip, width=2.3cm]{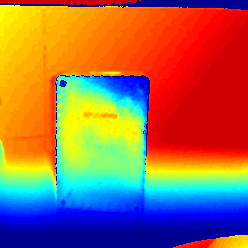}}
  \centerline{\includegraphics[trim=80px 15px 20px 35px, clip, width=2.3cm]{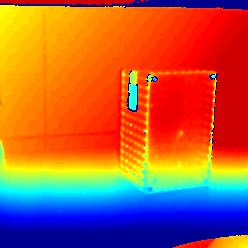}}
  \centerline{\includegraphics[trim=100px 40px 0px 10px, clip, width=2.3cm]{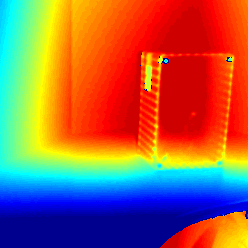}}
  \centerline{\includegraphics[trim=90px 50px 10px 10px, clip, width=2.3cm]{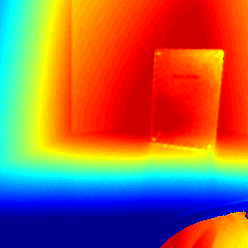}}
\end{minipage}
\hspace*{\fill}
\begin{minipage}[b]{.16\linewidth}
  \centering
  \centerline{(b) GT}\medskip
  \centerline{\includegraphics[trim=30px 10px 70px 40px, clip, width=2.3cm]{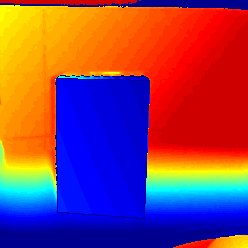}}
  \centerline{\includegraphics[trim=80px 15px 20px 35px, clip, width=2.3cm]{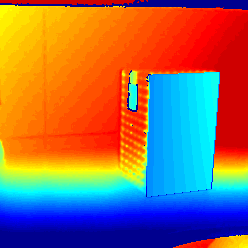}}
  \centerline{\includegraphics[trim=100px 40px 0px 10px, clip, width=2.3cm]{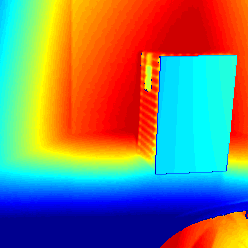}}
  \centerline{\includegraphics[trim=90px 50px 10px 10px, clip, width=2.3cm]{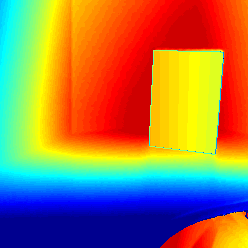}}
\end{minipage}
\hspace*{\fill}
\begin{minipage}[b]{.16\linewidth}
  \centering
  \centerline{(c) Shim \& Lee. \cite{shim2016recovering}}\medskip
  \centerline{\includegraphics[trim=30px 10px 70px 40px, clip, width=2.3cm]{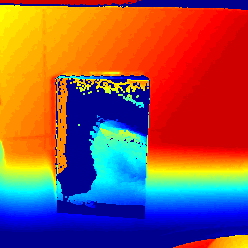}}
  \centerline{\includegraphics[trim=80px 15px 20px 35px, clip, width=2.3cm]{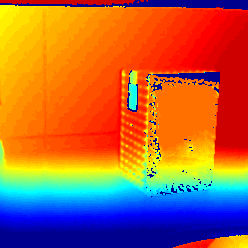}}
  \centerline{\includegraphics[trim=100px 40px 0px 10px, clip, width=2.3cm]{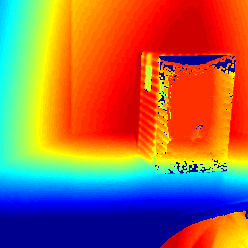}}
  \centerline{\includegraphics[trim=90px 50px 10px 10px, clip, width=2.3cm]{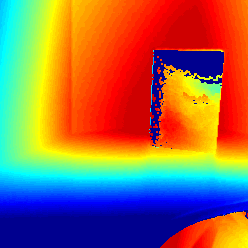}}
\end{minipage}
\hspace*{\fill}
\begin{minipage}[b]{.16\linewidth}
  \centering
  \centerline{(d) Ours}\medskip
  \centerline{\includegraphics[trim=30px 10px 70px 40px, clip, width=2.3cm]{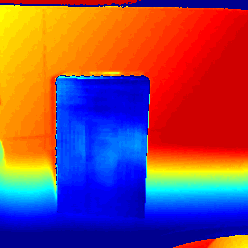}}
  \centerline{\includegraphics[trim=80px 15px 20px 35px, clip, width=2.3cm]{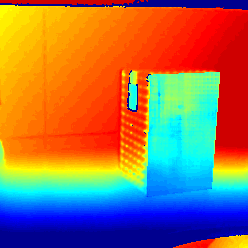}}
  \centerline{\includegraphics[trim=100px 40px 0px 10px, clip, width=2.3cm]{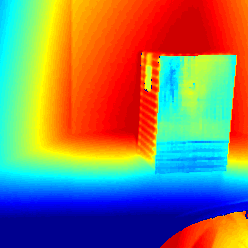}}
  \centerline{\includegraphics[trim=90px 50px 10px 10px, clip, width=2.3cm]{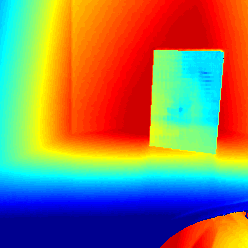}}
\end{minipage}
\hspace*{\fill}
\begin{minipage}[b]{.16\linewidth}
  \centering
  \centerline{(e) Ours filtered}\medskip
  \centerline{\includegraphics[trim=30px 10px 70px 40px, clip, width=2.3cm]{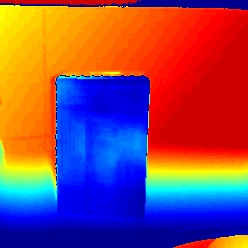}}
  \centerline{\includegraphics[trim=80px 15px 20px 35px, clip, width=2.3cm]{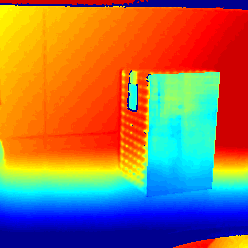}}
  \centerline{\includegraphics[trim=100px 40px 0px 10px, clip, width=2.3cm]{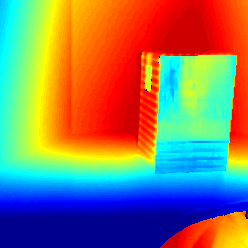}}
  \centerline{\includegraphics[trim=90px 50px 10px 10px, clip, width=2.3cm]{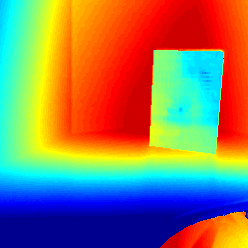}}
\end{minipage}
\hspace*{\fill}

\flushright{\includegraphics[width=4.2cm]{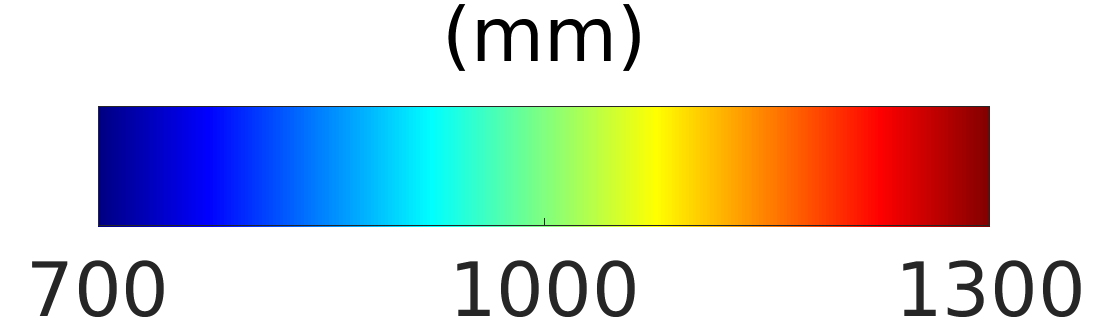}}

\caption{Qualitative comparisons on depth map estimation. Depth values outside the target object are identical to the raw depth values. Blue color indicates the pixel is close to the camera, as notated at bottom right}
\label{fig3}
\end{figure*}
\begin{figure}[tb]

\begin{minipage}[b]{.5\linewidth}
\flushleft{\includegraphics[width=4.2cm]{fig3/colorbar1.png}}
\vspace{0.3cm}
\end{minipage}
\begin{minipage}[b]{.5\linewidth}
\flushright{\includegraphics[width=1.9cm]{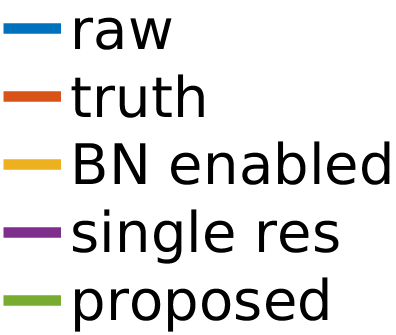}}
\end{minipage}%

\fboxsep=0mm
\fboxrule=2pt
\hspace*{\fill}
\begin{minipage}[b]{.16\linewidth}
  \centering
  \centerline{\fcolorbox{Dandelion}{white}{\includegraphics[trim=20px 25px 80px 25px, clip, width=2.2cm]{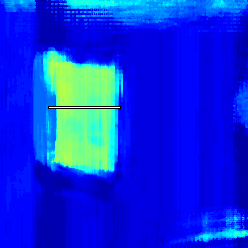}}}
  \vspace{0.3cm}
  \centerline{\fcolorbox{Dandelion}{white}{\includegraphics[trim=100px 30px 00px 20px, clip, width=2.2cm]{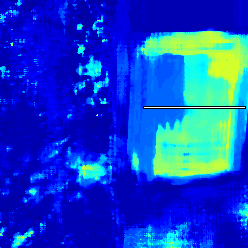}}}
  \vspace{0.3cm}\centerline{(a)}
\end{minipage}
\hspace*{\fill}
\begin{minipage}[b]{.16\linewidth}
  \centering
  \centerline{\fcolorbox{Plum}{white}{\includegraphics[trim=20px 25px 80px 25px, clip, width=2.2cm]{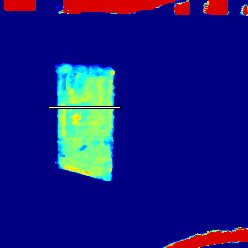}}}
  \vspace{0.3cm}
  \centerline{\fcolorbox{Plum}{white}{\includegraphics[trim=100px 30px 00px 20px, clip, width=2.2cm]{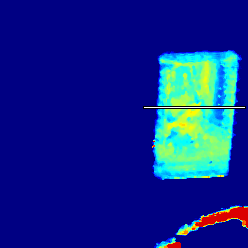}}}
  \vspace{0.3cm}\centerline{(b)}
\end{minipage}
\hspace*{\fill}
\begin{minipage}[b]{.16\linewidth}
  \centering
  \centerline{\fcolorbox{OliveGreen}{white}{\includegraphics[trim=20px 25px 80px 25px, clip, width=2.2cm]{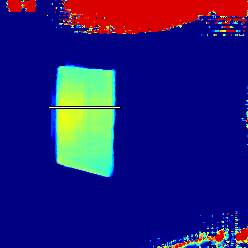}}}
  \vspace{0.3cm}
  \centerline{\fcolorbox{OliveGreen}{white}{\includegraphics[trim=100px 30px 00px 20px, clip, width=2.2cm]{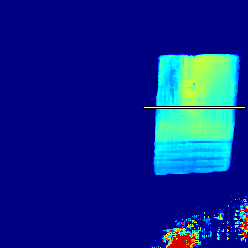}}}
  \vspace{0.3cm}\centerline{(c)}
\end{minipage}
\hspace*{\fill}
\begin{minipage}[b]{.35\linewidth}
  \centering
  \centerline{\includegraphics[width=4.5cm]{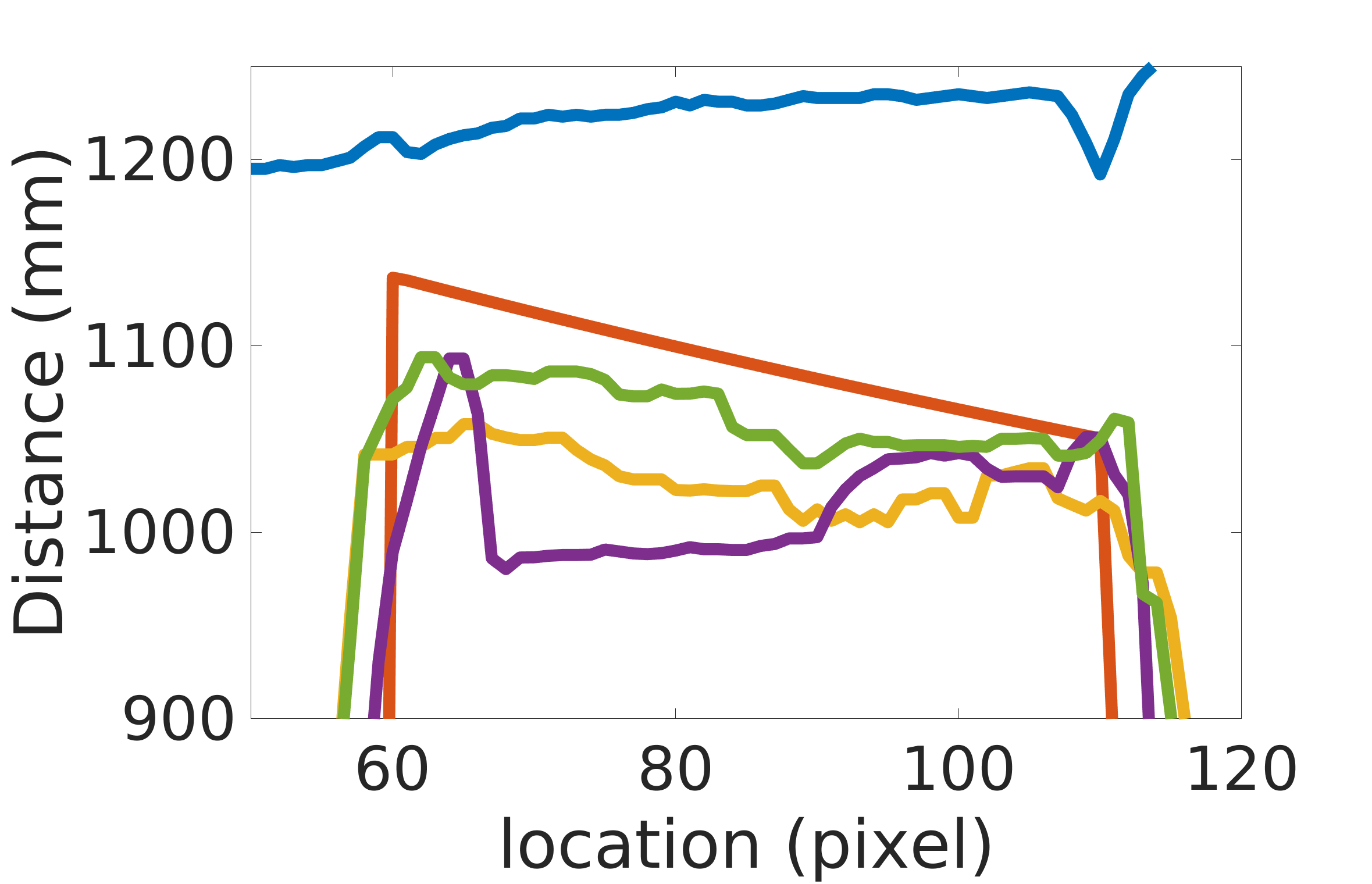}}
  \vspace{0.6cm}
  \centerline{\includegraphics[width=4.5cm]{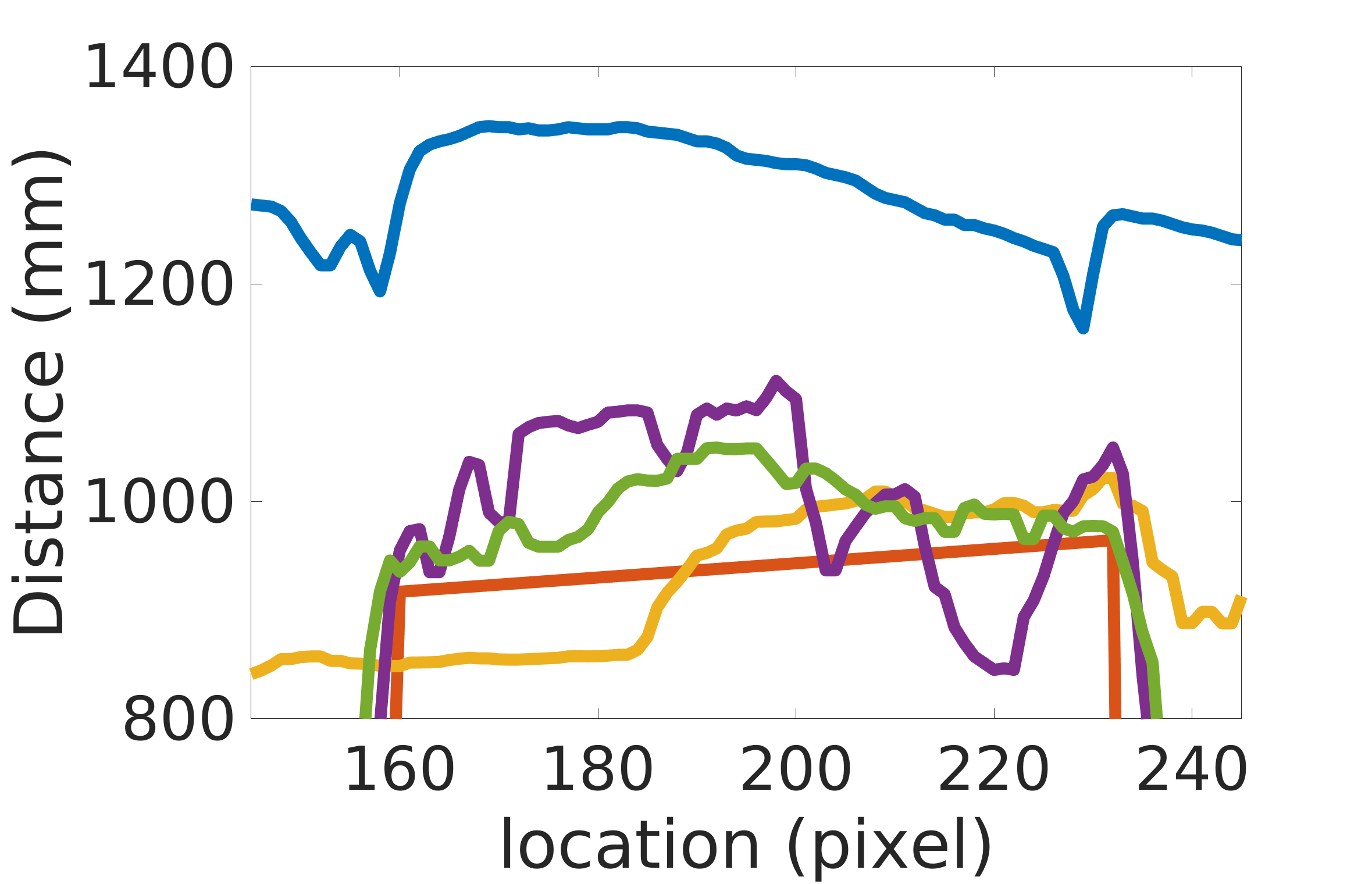}}
  \vspace{0.3cm}\centerline{\hphantom{000000}(d)}
\end{minipage}
\hspace*{\fill}

    \caption{Comparisons of different network architectures. (a) Network with BN layers, (b) network using the single scale patch, (c) proposed network, (d) center cut for visualizing the shape estimation performance (1-D plots correspond to the depth values along the white line of the images)}
    \label{fig4}
\end{figure}
\begin{table}[]
\centering
\caption{Accuracy comparison with previous work \cite{shim2016recovering} and different network architectures.  The best and second best results are boldfaced.}
\medskip
\setlength{\tabcolsep}{4pt}
\label{table1}
\begin{tabular}{@{}lrrrrrr@{}}
\toprule
                                        & \multicolumn{3}{c}{Planar object} & \multicolumn{3}{c}{Round objects} \\ \midrule
                                        & rms   & Rel   & $\log_{10}$  & rms   & Rel   & $\log_{10}$    \\ \midrule
raw data                                & 203.9 &  0.21 & 0.080        & 165.9 &  0.16 & 0.062          \\
Shim \& Lee. \cite{shim2016recovering}  & 177.9 &  0.18 & 0.069        & 147.4 &  0.13 & 0.053          \\ \midrule
batch normalization                     &  82.7 &  0.08 & 0.034        & 154.4 &  0.14 & 0.067          \\
single scale                       &  84.3 &  0.07 & 0.033        & \textbf{87.6} & \textbf{0.07} & \textbf{0.032} \\ \midrule
proposed                                & \textbf{70.4} & \textbf{0.06} & \textbf{0.028}       &  89.1 & \textbf{0.07} & \textbf{0.033} \\
proposed(+median filter)                & \textbf{70.2} & \textbf{0.06} & \textbf{0.028} & \textbf{89.0} & \textbf{0.07} & \textbf{0.033} \\ \bottomrule
\end{tabular}
\end{table}

\subsection{Quantitative and Qualitative Evaluation}

To quantitatively evaluate our results, following metrics are employed for measuring the depth errors.
\begin{itemize}
    \item Root mean squared error(rms): $\sqrt{\frac{1}{|T|} \sum_{d \in T} (\hat{d}-d )^2 }$ \hfill\eqnum\label{eqn1}
    \item Mean relative error(Rel): $\frac{1}{|T|} \sum_{d \in T} | \hat{d}-d | / d$ \hfill\eqnum\label{eqn2}
    \item Mean $\log_{10}$ error($\log_{10}$): $\frac{1}{|T|} \sum_{d \in T} |\log_{10}\hat{d} - \log_{10}d |$ \hfill\eqnum\label{eqn3}
\end{itemize}
$\hat{d}$ and $d$ means the predicted and ground-truth depth value, and $T$ denotes the set of translucent pixels. The pixels that are missing depth in raw data are excluded when computing $T$.

For qualitative evaluation, we demonstrate our results in Fig.~\ref{fig3}. Depth distortions by translucent objects are clear as seen by comparing Fig.~\ref{fig3}(a) and \ref{fig3}(b). We compare our results with those of Shim and Lee \cite{shim2016recovering}. This model ignores refraction, thereby is not suitable for handling the slanted surfaces. Consequently, the effect of depth correction is unclear in the second and third case of Fig.~\ref{fig3}(c). Meanwhile, our model drastically improves distorted depth maps regardless of surface orientations. Moreover, Shim and Lee's model produces noisy depth maps, thus post-processing throughout multiple exposure is critical as mentioned in their paper. On the contrary, our method performs consistent depth recovery even without any post-processing, as quantitatively stated in Table 1. Still, minor depth fluctuation can be observed because our network does not employ any external prior such as the refractive index, ground depth, or smoothness constraint.

We also show the effectiveness of our ideas for depth reconstruction. The qualitative and quantitative comparison with two degenerated networks are shown in Fig.~\ref{fig4}, and left half of Table~\ref{table1}, respectively. The network with a single-scale input (i.e., only a full-scale patch) has smaller receptive field, thereby the region far from edge has no visual clue of relative object structure and produces incorrect depth values, resulting in wiggly reconstructed surfaces. Another network with batch normalization (BN) layers tends to normalize each test batch (i.e., in our implementation, each vertical line forms the batch.). As a result, the recovered depth map exhibits the line-like artifacts as well as significant errors due to depth normalization. The proposed network, on the other hand, suffers from neither artifacts nor normalization error, and outperforms both networks quantitatively and qualitatively as shown in Table~\ref{table1}.

\begin{figure*}[!t]

\hspace*{\fill}
\begin{minipage}[b]{.14\linewidth}
  \centering
  \centerline{(a) Raw depth}\medskip
  \centerline{\includegraphics[trim=120px 50px 40px 100px, clip, width=2.0cm]{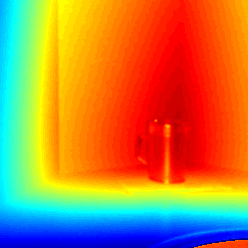}}
  \centerline{\includegraphics[trim=130px 30px 10px 90px, clip, width=2.0cm]{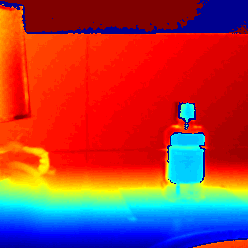}}
  \centerline{\includegraphics[trim=100px 10px 0px 70px, clip, width=2.0cm]{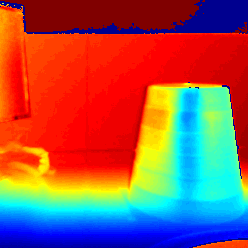}}
\end{minipage}
\hspace*{\fill}
\begin{minipage}[b]{.14\linewidth}
  \centering
  \centerline{(b) GT}\medskip
  \centerline{\includegraphics[trim=120px 50px 40px 100px, clip, width=2.0cm]{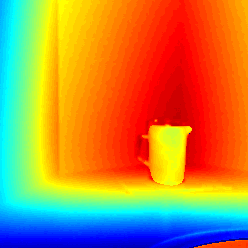}}
  \centerline{\includegraphics[trim=130px 30px 10px 90px, clip, width=2.0cm]{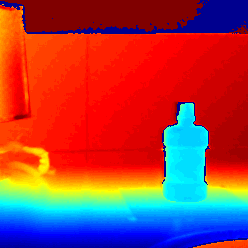}}
  \centerline{\includegraphics[trim=100px 10px 0px 70px, clip, width=2.0cm]{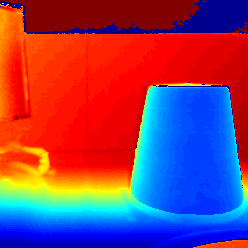}}
\end{minipage}
\hspace*{\fill}
\begin{minipage}[b]{.14\linewidth}
  \centering
  \centerline{(c) Shim \& Lee.}\medskip
  \centerline{\includegraphics[trim=120px 50px 40px 100px, clip, width=2.0cm]{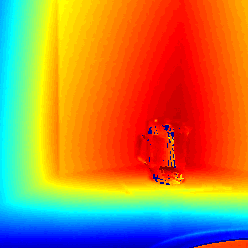}}
  \centerline{\includegraphics[trim=130px 30px 10px 90px, clip, width=2.0cm]{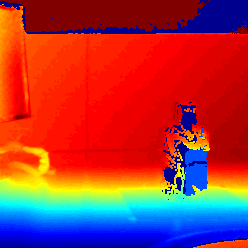}}
  \centerline{\includegraphics[trim=100px 10px 0px 70px, clip, width=2.0cm]{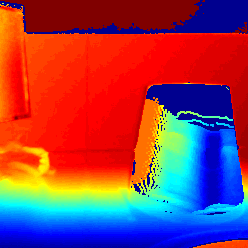}}
\end{minipage}
\hspace*{\fill}
\begin{minipage}[b]{.14\linewidth}
  \centering
  \centerline{(d) BN}\medskip
  \centerline{\includegraphics[trim=120px 50px 40px 100px, clip, width=2.0cm]{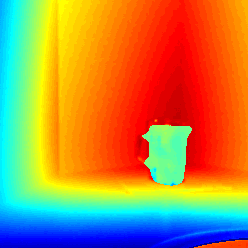}}
  \centerline{\includegraphics[trim=130px 30px 10px 90px, clip, width=2.0cm]{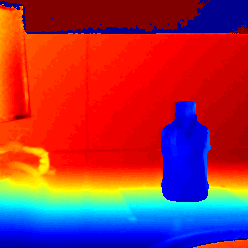}}
  \centerline{\includegraphics[trim=100px 10px 0px 70px, clip, width=2.0cm]{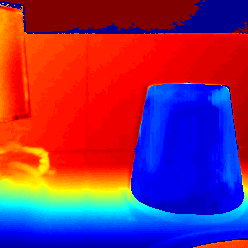}}
\end{minipage}
\hspace*{\fill}
\begin{minipage}[b]{.14\linewidth}
  \centering
  \centerline{(e) 1-res}\medskip
  \centerline{\includegraphics[trim=120px 50px 40px 100px, clip, width=2.0cm]{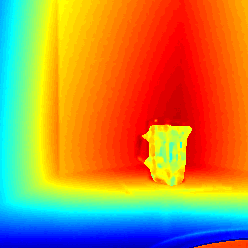}}
  \centerline{\includegraphics[trim=130px 30px 10px 90px, clip, width=2.0cm]{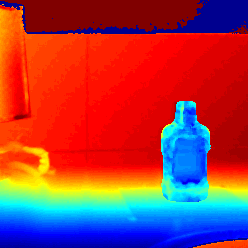}}
  \centerline{\includegraphics[trim=100px 10px 0px 70px, clip, width=2.0cm]{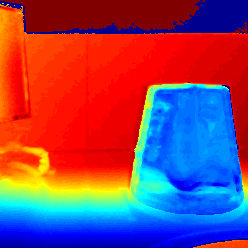}}
\end{minipage}
\hspace*{\fill}
\begin{minipage}[b]{.14\linewidth}
  \centering
  \centerline{(f) Ours filtered}\medskip
  \centerline{\includegraphics[trim=120px 50px 40px 100px, clip, width=2.0cm]{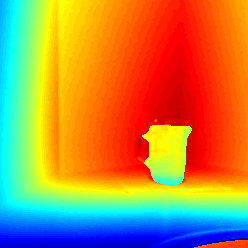}}
  \centerline{\includegraphics[trim=130px 30px 10px 90px, clip, width=2.0cm]{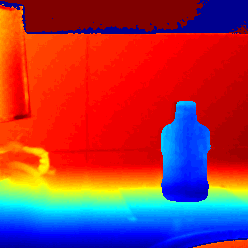}}
  \centerline{\includegraphics[trim=100px 10px 0px 70px, clip, width=2.0cm]{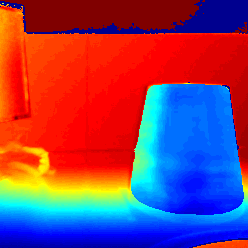}}
\end{minipage}
\hspace*{\fill}

\flushright{\includegraphics[width=4.0cm]{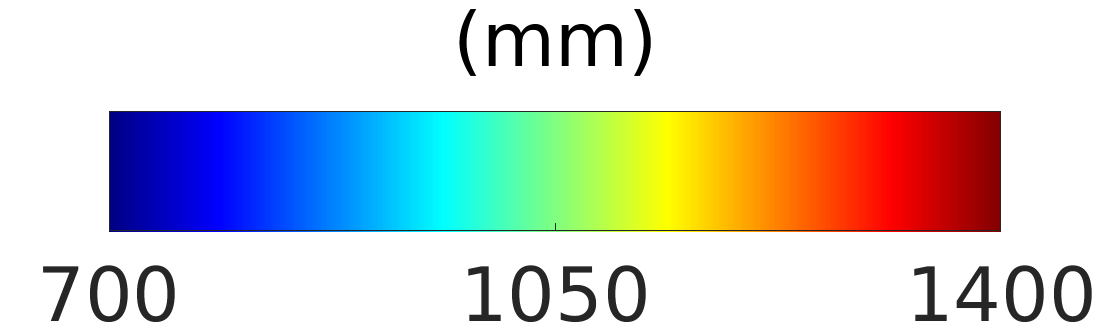}}

\caption{Depth estimations using round objects. Depth values outside the object is identical to the raw depth values. `BN' and `1-res' stand for each of two degraded networks, respectively; the network with BN layers and the single-resolution input network. Yet, none of the models are trained on round objects}
\label{fig5}
\end{figure*}

We further evaluate our model using round objects while the proposed model is trained with solely planar objects. The proposed model reports reasonable performance, even though the objects exhibit completely different characteristics from training dataset, in terms of both 2D outlines and 3D structures. Interestingly, as shown in the right half of Table 2, the network with single-scale patch input demonstrates marginally better quantitative results than the proposed network for recovering the round objects. In fact, this is expected because the single-scale network focuses on modeling the local surfaces, equivalent to a small receptive field. At the local scale, the curved surface can be reasonably approximated by the piece-wise linear surface. 
Contrarily, as shown as the qualitative comparisons presented in Fig.~\ref{fig5}, the output of the single-scale patch network presents small ripples as artifacts in the estimated depth maps. These results are analogous to those of planar objects. The proposed network generally produces smooth surfaces. Also, our model handles the object with less curved surfaces better; for example, the depth estimation of \emph{Garbage Bin} (Fig.~\ref{fig5}, bottom row) is more accurate than others, because this object is larger in size and less curved than others.

\subsection{Noise Robustness}

We analyze our performance by increasing the level of input noise. This empirical study is formulated to show how much the proposed model is robust against the input noise. In fact, the noise resiliency is a critical property for processing the measurements from depth cameras. It is because 1) active depth sensors always suffer from the lack of emitted light energy, and 2) the short exposure is necessary for reducing the motion blurs, thus inevitable for increasing the temporal noise. 

To simulate the sensor noise, we refer the empirical noise model by Belhedi et al \cite{belhedi2015noise}, thus choose a distance-independent Gaussian distribution as the noise distribution for ToF sensors. 

\begin{table}[!t]
\centering
\caption{Performance of the proposed model by increasing the standard deviation ($\sigma$) of additive white Gaussian noise. Note that no postprocessing is applied}
\medskip
\setlength{\tabcolsep}{4pt}
\label{table2}
\begin{tabular}{@{}l|rrr@{}}
\toprule
\multicolumn{1}{c|}{$\sigma$ (mm)} & \multicolumn{1}{c}{rms} & \multicolumn{1}{c}{Rel} & \multicolumn{1}{c}{$\log_{10}$} \\ \midrule
0 & 70.4 & 0.06 & 0.028 \\
0.5 & 70.4 & 0.06 & 0.028 \\
1 & 70.3 & 0.06 & 0.028 \\
2 & 70.5 & 0.06 & 0.028 \\
4 & 70.4 & 0.06 & 0.028 \\
8 & 70.4 & 0.06 & 0.028 \\
16 & 69.4 & 0.06 & 0.027 \\
32 & 72.8 & 0.07 & 0.030 \\
64 & 102.5 & 0.09 & 0.044 \\
128 & 143.8 & 0.13 & 0.063 \\ \bottomrule
\end{tabular}
\end{table}
\begin{figure*}[t]

\hspace*{\fill}
\begin{minipage}[b]{.16\linewidth}
  \centering
  \centerline{(a) GT}\medskip
  \centerline{\includegraphics[trim=10px 20px 90px 30px, clip, width=2.3cm]{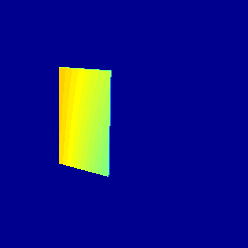}}
\end{minipage}
\hspace*{\fill}
\begin{minipage}[b]{.16\linewidth}
  \centering
  \centerline{(b) $\sigma=0$mm}\medskip
  \centerline{\includegraphics[trim=10px 20px 90px 30px, clip, width=2.3cm]{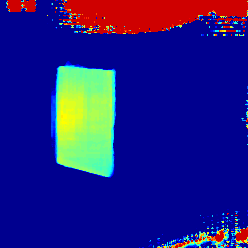}}
\end{minipage}
\hspace*{\fill}
\begin{minipage}[b]{.16\linewidth}
  \centering
  \centerline{(c) $\sigma=8$mm}\medskip
  \centerline{\includegraphics[trim=10px 20px 90px 30px, clip, width=2.3cm]{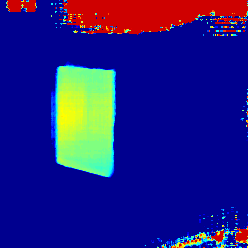}}
\end{minipage}
\hspace*{\fill}
\begin{minipage}[b]{.16\linewidth}
  \centering
  \centerline{(d) $\sigma=32$mm}\medskip
  \centerline{\includegraphics[trim=10px 20px 90px 30px, clip, width=2.3cm]{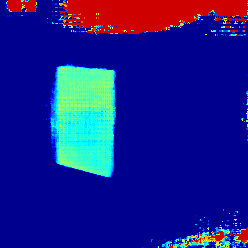}}
\end{minipage}
\hspace*{\fill}
\begin{minipage}[b]{.16\linewidth}
  \centering
  \centerline{(e) $\sigma=128$mm}\medskip
  \centerline{\includegraphics[trim=10px 20px 90px 30px, clip, width=2.3cm]{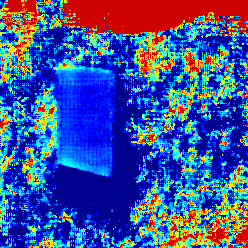}}
\end{minipage}
\hspace*{\fill}

\flushright{\includegraphics[width=3.6cm]{fig3/colorbar1.png}}

\caption{Qualitative performance of depth estimation upon various noise levels}
\label{fig6}
\end{figure*}

Table~\ref{table2} and Figure~\ref{fig6} summarizes quantitative and qualitative results of our model under various levels of input noise. Fortunately, the noise with standard deviation of $16 mm$ or less does not much degrade our performance. In fact, Sarboland et al. \cite{sarbolandi2015kinect} report that, except for near-corner pixels, it is unusual for the Tof depth sensor (Kinect v2) to suffer from sensor noise greater than the standard deviation of $20 mm$. Therefore, it is reasonable to claim that the proposed model is robust against the shot noise produced by the ToF sensors. In case of the extreme noise variations (after $32 mm$), our model starts to malfunction in terms of both overall depth estimation and minor artifacts as seen from Fig.~\ref{fig6}(d)-(e).


\section{Conclusions}

We present a deep residual network architecture for recovering depth distortion from translucent object, using a single time-of-flight (ToF) depth camera. We propose the multi-scale patch representation and exclusion of batch normalization for developing the network architecture, and show that they are effective to improve the accuracy of reconstructing translucent objects. The quantitative and qualitative evaluations over the competitor clearly demonstrate the superiority of our model; we report higher accuracy and show the robust performance for handling various object poses and optical properties. In addition, the experimental validation of our proposals justifies their positive effects. By showing the robustness of our model across various levels of input noise, we highlight that our model can be a practical solution for real applications. To the best of our knowledge, we propose the first approach to recovering the 3-D translucent object using deep neural networks. We hope that our work can initiate learning based approaches to the problem of recovering translucent objects. In the future, we plan to expand our idea to the large scale dataset, which covers a wide range and shape of translucent objects.

\bibliographystyle{splncs}
\bibliography{egbib}

\end{document}